\documentclass[letterpaper]{article} 
\usepackage{aaai23}  
\usepackage{times}  
\usepackage{helvet}  
\usepackage{courier}  
\usepackage[hyphens]{url}  
\usepackage{graphicx} 
\usepackage{natbib}  
\usepackage{caption} 
\usepackage{stfloats}
\usepackage[linesnumbered,ruled,vlined]{algorithm2e}
\usepackage{multirow}
\usepackage{amsmath}
\usepackage{booktabs}
\usepackage{tablefootnote}
\usepackage{float}

\usepackage{color}

\title{AAAI Press Anonymous Submission\\Instructions for Authors Using \LaTeX{}}
\author{
    Qianyu He\textsuperscript{\rm 1}, 
    Xintao Wang\textsuperscript{\rm 1},
    Jiaqing Liang\textsuperscript{\rm 2}\thanks{Corresponding author.},
    Yanghua Xiao\textsuperscript{\rm 1,3}\footnotemark[1]
}
\affiliations{
    \textsuperscript{\rm 1}Shanghai Key Laboratory of Data Science, School of Computer Science, Fudan University\\
    \textsuperscript{\rm 2}School of Data Science, Fudan University\\
     \textsuperscript{\rm 3}Fudan-Aishu Cognitive Intelligence Joint Research Center, Shanghai, China\\

    \{qyhe21, xtwang21\}@m.fudan.edu.cn, liangjiaqing@fudan.edu.cn, shawyh@fudan.edu.cn
}

\title{MAPS-KB: A Million-scale Probabilistic Simile Knowledge Base}

\usepackage{bibentry}

\begin{document}

\maketitle

\begin{abstract}

The ability to understand and generate similes is an imperative step to realize human-level AI. 
However, there is still a considerable gap between machine intelligence and human cognition in similes, since deep models based on statistical distribution tend to favour high-frequency similes.
Hence, a large-scale symbolic knowledge base of similes is required, as it contributes to the modeling of diverse yet unpopular similes while facilitating additional evaluation and reasoning.
To bridge the gap, we propose a novel framework for large-scale simile knowledge base construction, as well as two probabilistic metrics which enable an improved understanding of simile phenomena in natural language.
Overall, we construct MAPS-KB, a million-scale probabilistic simile knowledge base, covering 4.3 million triplets over 0.4 million terms from 70 GB corpora.
We conduct sufficient experiments to justify the effectiveness of methods of our framework. We also apply MAPS-KB on three downstream tasks to achieve state-of-the-art performance, further
demonstrating the value of MAPS-KB.
Resources of MAPS-KB are publicly available at \url{https://github.com/Abbey4799/MAPS-KB}.


\end{abstract}

\section{Introduction}

As a figurative language, metaphors allow people to understand abstract concepts through concrete and familiar ones~\cite{lakoff1993contemporary, lakoff2008metaphors}.
Metaphor understanding has become one of the most fundamental tasks to evaluate cognition intelligence~\cite{liu2022testing,chen2022probing}.
A simile is a special type of metaphor, which compares two fundamentally different terms via shared properties~\cite{paul1970figurative,tversky1977features}. 
For example, in simile ``\textit{Her hair felt like silk.}'', ``\textit{hair}'' (a.k.a, \textit{\emph{topic}}) is compared with ``\textit{silk}'' (a.k.a., \textit{\emph{vehicle}}) with the implicit \textit{\emph{property}} ``\textit{soft}'', where \textit{topic}, \textit{vehicle} and \textit{property} are three main components of similes~\cite{hanks2013lexical}.
Since similes can make the literal expression more imaginative and graspable, they have been widely used in literature~\cite{fishelov2007shall} and daily conversations~\cite{niculae2014brighter}. 


Nowadays, large pre-trained language models (PLMs) have become an important foundation of human-level machine intelligence.
PLMs have achieved state-of-the-art performance in various language processing tasks.
However, PLMs still lag behind human cognition.
For example, while PLMs exhibit a certain ability to interpret similes after fine-tuning~\cite{chakrabarty2021s, he2022can}, they can hardly understand similes that require common-sense knowledge~\cite{liu2022testing}.
Although PLMs can do polish writing by generating coherent similes~\cite{zhang2020writing, chakrabarty2020generating}, the generated similes are far less diverse than human-written ones, since statistical models tend to favor high-frequency similes.

A large-scale symbolic knowledge base of similes in general is needed to further improve the simile understanding ability of PLMs.
Symbolic knowledge bases have been widely introduced to bridge the gap between machines and human intelligence~\cite{liu2020kbert}.
Firstly, PLMs and other statistical models themselves suffer from a lack of knowledge that rarely appears in the corpora.
Besides, \emph{explicit structured} knowledge is indispensable for many applications that require explanation and interpretability, while PLMs only contain \emph{implicit statistical} knowledge that is hard to understand. 
Furthermore, a structured knowledge base can provide additional analytical conveniences compared with purely statistical models, such as taxonomy-based reasoning and path-based reasoning.
Hence, a large-scale simile knowledge base can help machines understand the diverse yet rarely expressed similes.

Many efforts have been devoted to the construction of large-scale symbolic knowledge bases,
(KBs, or knowledge graphs)
such as entity-oriented KBs~\cite{vrandevcic2014wikidata}, concept-oriented KBs~\cite{wu2012probase}, and word-oriented KBs~\cite{miller1995wordnet}.
Recently, there has also been an increasing number of simile knowledge resources~\cite{roncero2015semantic, liu2022testing,li2013data}.
However, they are still unsatisfactory, concerning both their coverage and expressiveness.
On one hand, most of them are only thousand-scale~\cite{roncero2015semantic, liu2022testing},
while million-scale knowledge bases are needed in a wide range of downstream applications, such as online recommendation systems~\cite{wang2019kgat} and open-domain question answering~\cite{cui2019kbqa}. 
On the other hand, existing simile knowledge bases skip acquisition of \textit{property}~\cite{li2013data}, an important component of similes, since \textit{properties} are usually implicit and can not be extracted from texts directly, such as the implicit property ``\textit{soft}'' in  ``\textit{Her hair felt like silk.}''.
However, the shared \textit{property} is the foundation for comparisons between the \textit{topic} and \textit{vehicle}, and is thus essential in  downstream tasks~\cite{chakrabarty2020generating}.

\begin{figure*}[htbp]
    \centering
        \includegraphics[width=0.9\linewidth]{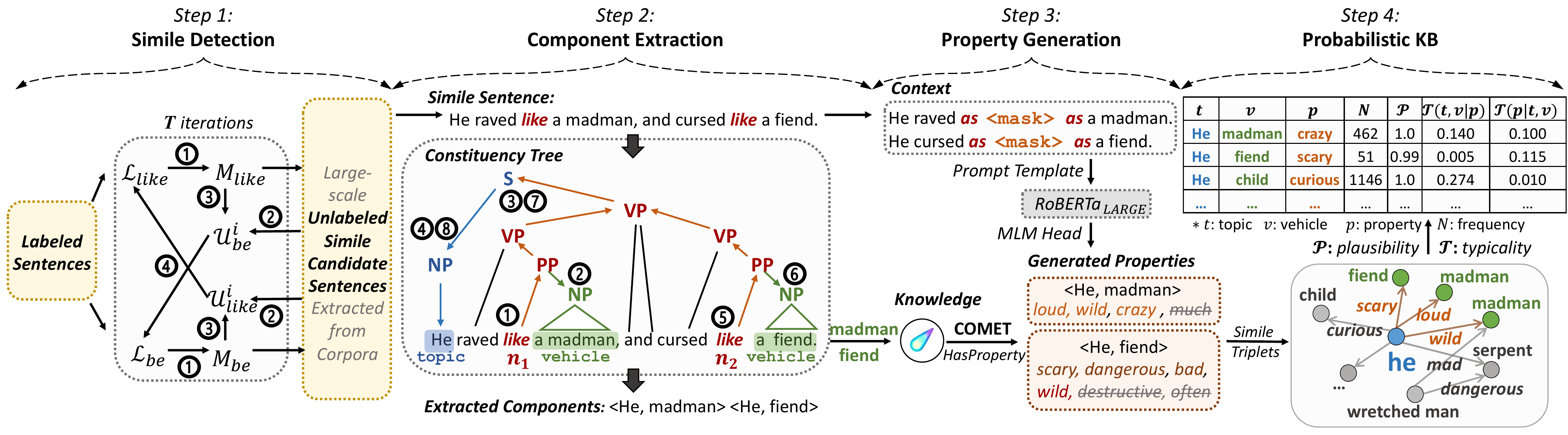} 
    \captionsetup{font={small}} 
    \caption{Our framework for large-scale simile knowledge base construction. 
    In Step 1, with large-scale corpora and a small set of labeled sentences as input, we detect millions of simile sentences via \textit{co-training}.
    In Step 2, given each simile sentence, we extract the important component \textit{topic} and \textit{vehicle} by designing rules based on constituency parsing. 
    In Step 3, we generate the important component \textit{property} considering both \textit{context} information and external \textit{knowledge} sources, and finally complete millions of simile triplets $(t,p,v)$.
    In Step 4, we design two probabilistic metrics (\textit{plausibility} and \textit{typicality}) for each simile triplet, which can facilitate simile-related inference.}
    \label{fig:framework}
\end{figure*}

In this paper, we construct \textbf{MAPS-KB}, a \textbf{M}illion-sc\textbf{A}le \textbf{P}robabilistic \textbf{S}imile \textbf{K}nowledge \textbf{B}ase.
The construction framework is shown in Figure \ref{fig:framework}.
In the first step, we extract simile candidate sentences via two syntactic patterns, then distinguish simile sentences from literal sentences through \textit{co-training}, leveraging large-scale unlabeled sentences collected from both patterns.
In the second step, we extract the simile components (\textit{topic} and \textit{vhicle}) by designing rules based on constituency parsing, addressing two challenges: \textit{simile component positioning} with long distance dependencies and \textit{simile component span identification} at the same time.
In the third step, we generate the simile component \textit{property} for each simile sentence, considering both surrounding \textit{context} information and external \textit{knowledge} sources, since properties can hardly be extracted from texts directly. 
Furthermore, we intentionally retain each simile triplet with its frequency information and design two probabilistic metrics for simile triplets in the fourth step, which allow machines to better understand similes in natural language and facilitate related applications in the real world.

To summarize, our contributions are mainly three-fold:
(1) To the best of our knowledge, MAPS-KB is the first million-scale probabilistic simile knowledge base, covering 4.3 million triplets over 0.4 million terms from 70 GB corpora.
The probabilistic metrics \textit{plausibility} and \textit{typicality} in MAPS-KB endow machines with an improved understanding of simile phenomena and facilitate a wide range of real applications.
(2) We propose a novel framework for large-scale probabilistic simile knowledge base construction.
This framework can be extended to other figurative languages like sarcasm and humor which can be represented by several major components.
(3) We conduct extensive experiments to evaluate the effectiveness and necessity of methods in our framework, and further use MAPS-KB to 
achieve state-of-the-art performance in three popular simile-related tasks. 



\section{Related Work}
\subsubsection{Simile Knowledge Acquisition}

Early studies mainly acquire simile knowledge via linguists~\cite{lakoff2008metaphors,roncero2015semantic}.
However, the acquisition process is human-laborious and inefficient.
With the rise of web documents, data-driven methods are developed to acquire simile knowledge automatically~\cite{li2013data}.
\cite{li2013data} first collect simile sentences via syntactic patterns, and then retain the topic-vehicle pairs without is-a relations to construct a million-scale simile knowledge base.
Recently, many studies probe simile knowledge from PLMs via designed prompt template~\cite{chen2022probing, he2022can}.
Different from these works, we automatically construct a million-scale probabilistic simile knowledge base in the form of simile triplets.

\subsubsection{Simile Processing Tasks}
Earlier studies mainly focus on discriminating similes from literal sentences and extracting simile components from similes~\cite{liu2018neural,zeng2020neural}.
Recently, many studies transfer literal sentences to similes via finetuning PLMs~\cite{chakrabarty2020generating,zhang2020writing}. 
There is also a bulk of works designing specific tasks to investigate whether PLMs have the ability to interpret similes, such as simile property probing task~\cite{he2022can}, Winograd-style simile understanding~\cite{liu2022testing} and simile continuation selection and generation~\cite{chakrabarty2021s}.
Different from these works, we perform simile processing tasks based on a large-scale probabilistic simile knowledge base.

\section{MAPS-KB Construction}

\begin{table}[h] 
    %
    \centering
    \resizebox{0.38\textwidth}{11mm}{
        \begin{tabular}{c|c|c|c}
        \toprule  
       \textbf{Name}        & \textbf{Size} & \textbf{\# $\text{Pattern}_{\text{like}}$} & \textbf{\# $\text{Pattern}_{\text{be}}$}  \\
        \midrule  
        Openwebtext &   38GB           &    1.5M     &         21M     \\
        Gutenburg   &   26GB          &     0.5M     &          8M        \\
        Bookcorpus  &   6GB            &    0.5M     &          2M       \\
        \midrule  
        overall   &      70GB        &  2M    &      31M           \\
        \bottomrule 
        \end{tabular}
    }
    
       \captionsetup{font={small}} 
        \caption{Statistics of used corpora and extracted simile candidates via two syntactic patterns. G and M mean Gigabytes and millions.}    
    \label{tab:corpus_statistics}
    
\end{table}

We start by collecting sentences from several corpora (Table~\ref{tab:corpus_statistics}).
Then, we design two syntactic patterns to extract sentences.
According to~\cite{paul1970figurative}, similes compare two different terms, typically using the comparator ``\textit{like}'', such as ``\textit{Her hair felt like silk}''.
Additionally, after removing the comparator ``\textit{like}'', similes become implicit comparisons~\cite{tversky1977features}, such as ``\textit{Her hair is silk.}''.
Considering these linguistic theories, our extraction patterns are defined as follows: (1) the \textit{like} pattern: \textit{$\text{Noun}_1$} ... BE/VB like ... \textit{$\text{Noun}_2$};  (2) the \textit{be} pattern: \textit{$\text{Noun}_1$} ... BE ... \textit{$\text{Noun}_2$}.
Finally, the extracted sentences are \textit{simile candidates}, which can either be true similes or literal comparisons.


\subsubsection{Simile Detection}

Next, we distinguish simile sentences from literal sentences.
There are two syntactic patterns of data in the collected unlabeled sentences, namely the \textit{like} pattern and the \textit{be} pattern.
Since expressions of two patterns for the same topic-vehicle pair are semantically consistent,
we adopt \textit{co-training} for simile detection.
The sentences extracted by the two patterns are seen as instances from two \textit{views}, namely \textit{like} view and \textit{be} view.

In a bootstrapping process of semi-supervised co-training, classifiers from multiple views are trained in a collaborative manner.
Compared with the self-training procedure which only utilizes data from one view, the co-training procedure makes models from \textit{like} view and \textit{be} view complement each other.
Compared with other supervised methods~\cite{zeng2020neural}, co-training is semi-supervised and effectively leverages massive unlabeled sentences.

Our detailed simile detection procedure with co-training is illustrated in Algorithm \ref{alg:simile_detection}.
At each iteration $i$, two separate classifiers $M_{like}$ and $M_{be}$ are trained by the data from \textit{like} view and \textit{be} view respectively (line 3-7).
Here, we use two seperate $\text{BERT}_{\text{BASE}}$ model as our classifiers.
Then, $M_{like}$ and $M_{be}$ annotate the unlabeled subset of the other's view  $\mathcal U^i_{be}$ and $\mathcal U^i_{like}$ (line 8-11).
Afterwards, the pseudo-labeled subsets $\mathcal U^i_{be}$ and $\mathcal U^i_{like}$ are integrated into the labeled set $\mathcal L_{be}$, $\mathcal L_{like}$ (line 13-14).
The process is performed iteratively, and there are $T$ iterations in total. 
After co-training, the fully trained classifiers $M_{like}$ and $M_{be}$ are applied to $\mathcal U_{be}$ and $\mathcal U_{like}$ again respectively. 
Positively labeled samples in final round are denoted as $\mathcal L'_{like}$ and $\mathcal L'_{be}$.

To improve the quality of pseudo-labels, only sentences with predicted confidence score greater than certain thresholds ($\theta_{like}$ and $\theta_{be}$) would be labeled as simile sentences.
Since there are much fewer positive samples than negative ones, pseudo-labeled negative samples are sampled to an equal number as positive ones to avoid data imbalance (line 12).
A small set of data from \textit{like} view are labeled to initialize the two classifiers at the first iteration.

Eventually, we extract 637,253 simile sentences from $like$ view and 1,279,537 simile sentences from $be$ view.

\begin{algorithm}[t]
    \DontPrintSemicolon
  \caption{Simile Detection with Co-training.} \label{alg:simile_detection}
  \begin{footnotesize}
  \KwIn{Labeleded  sentence set $\mathcal L_{like}$. Unlabeled sentence set $\mathcal U_{like}$, $\mathcal U_{be}$. Iteration times $T$. Sample ratio $\alpha_{like},\alpha_{be}$. Classifiers $M_{like}$, $M_{be}$.}
  \KwOut{Simile sentences $\mathcal L'_{like}$, $\mathcal L'_{be}$.}
  Initiate $\mathcal L_{be}$ with $\{\}$.\;
    \For{$i = 0 \to T$}{
        \eIf{$i == 0$}{
            Train $M_{like}$, $M_{be}$ with $\mathcal L_{like}$.\;}
          {
            Train $M_{like}$ with $\mathcal L_{like}$. \;
            Train $M_{be}$ with $\mathcal L_{be}$. \;
          }
    Sample $\alpha_{like}$ of $\mathcal U_{like}$ to get $\mathcal U^i_{like}$.\;
    Sample $\alpha_{be}$ of $\mathcal U_{be}$ to get $\mathcal U^i_{be}$.\;
    Use $M_{like}$ to label $\mathcal U^i_{be}$.\;
    Use $M_{be}$ to label $\mathcal U^i_{like}$.\;
    Balance the proportion of positive and negative samples in $\mathcal U^i_{be}$ and $\mathcal U^i_{like}$.\;
    Update $\mathcal L_{like}$ with pseudo-labeled $\mathcal U^i_{like}$.\;
    Update $\mathcal L_{be}$ with pseudo-labeled $\mathcal U^i_{be}$.\;
    }
Label $\mathcal U_{lbe}$ by $M_{like}$ to get $\mathcal L'_{be}$.\;
Label $\mathcal U_{like}$ by $M_{be}$ to get $\mathcal L'_{like}$.\;
\KwRet $\mathcal L'_{like}$, $\mathcal L'_{be}$.
    \end{footnotesize}
  \end{algorithm}

\subsubsection{Components Extraction}



In this section, we extract the simile components \textit{topic} $t$ and \textit{vehicle} $v$ from simile sentences.
There are two main challenges: \textit{component positioning} and \textit{component span identification}.
Take the simile ``\textit{They were like kids in a candy store.}'' as an example.
Component positioning is to locate the object ``\textit{kid}'' which the topic ``\textit{They}'' is compared to.
Given the object ``\textit{kid}'', component span identification determines the optimal span for the semantics that the vehicle should carry.
As  a vehicle, ```\textit{kids in the candy store}'' is more expressive than ``\textit{kids}'' to represent how enthusiastic ``\textit{They}'' were.
Hence, the method not only needs to locate the components accurately, but also needs to detect suitable component spans.

To address these two challenges, we design rules based on \textit{constituency parsing} of simile sentences.
Previous studies~\cite{liu2018neural,zeng2020neural} ask annotators to label the training dataset and train the sequence labeling models.
However, the models may perform worse for sentences with long-distance dependencies.
Constituency parsing parses sentences into {trees of} sub-phrases.
As illustrated in the second step in Figure \ref{fig:framework}, the leaf nodes are words, while the non-leaf nodes are the types of phrases in a constituency tree.
The phrase-level parsing allows constituency parsing to address both \textit{component positioning} with long distance dependencies and \textit{component span identification}.

The pseudo-code of our component extraction is illustrated in Algorithm \ref{alg:component_extraction}$\footnote{The proposed method is based on similes from the ``\textit{like}'' pattern. 
Extracting components from similes of the ``be'' pattern is left for future work.
Because our main focus is to construct a million-scale simile KB and prioritize its quality.
As for the coverage, ~\cite{niculae2014brighter} found that 82\% similes contain the comparator ``like" among the collected datasets. 
As for the precision, the ``like” pattern similes can be better extracted with less noise, due to explicit structure. }$.
First, we find all the leaf nodes $n_i$ whose text is ``\textit{like}'' as anchor nodes (line 10).
This is because the subject and object of the comparator ``\textit{like}'' are usually the \textit{topic} and \textit{vehicle}.
According to our observation, the \textit{vehicle} generally appears in the subtree rooted at $n_i$'s sibling node (the green part) (line 12-13).
Moreover, we iterate over the parent nodes of $n_i$ from bottom to the top (the red part) until a node labeled as \texttt{S} or \texttt{NP} (line 15-20).
The \textit{topic} often appears in the subtree rooted at this node (the blue part).
Second, we use the function GETCOMP for located subtrees to find the component.
The function GETCOMP first performs pre-order traversal over the subtree, then returns the first node whose child nodes are all labeled by \texttt{NP} or \texttt{PP} (line 2-8).
Even if a node is labeled by \texttt{NP}, its text may be a complex clause.
Hence, ignoring the labels of child nodes may extract improper components.
Finally, we further filter simile sentences based on extracted (topic, vehicle) pairs, due to the noise in the simile detection process.
Some special cases for component extraction and the filter rules are detailed in Appendix.
Overall, we collect 524,055 simile sentences with extracted topics and vehicles from 637,253 \textit{like}-view sentences.




\begin{algorithm}[t]
   \caption{Component Extraction Based on Constituency Parsing.}
   \label{alg:component_extraction}
   \begin{footnotesize}
   \KwIn{Parsed constituency tree $\mathcal K$ of sentence $S$.}
   \KwOut{Extracted component topic $c_t$ and vehicle $c_v$.}
   
  \DontPrintSemicolon
  \SetKwFunction{FMain}{GETCOMP}
  \SetKwProg{Fn}{Function}{:}{}
  \Fn{\FMain{Tree Node $n$}}{
  
        \While{$n \neq  \texttt{NULL}$}{
        $\mathcal{L} = $ \textit{label} set of $n \rightarrow \text{\textit{children}}$.\;
        \If{$\mathcal{L}$ are all in $\{\texttt{NP},\texttt{PP}\}$ \textbf{and} \texttt{NP} in $\mathcal{L}$}{
          \KwRet $n \rightarrow \text{\textit{text}}$.\;
          }
        \For{$c \in n \rightarrow \text{\textit{children}}$}{
        \If{\FMain{c} $\neq$ \texttt{NULL}}{
            \KwRet \FMain{c}.\;}
            }
        }
        \KwRet \texttt{NULL}.\;
  }
$\mathcal T = \{ n_0,n_1, ..., n_m\}$ is a subset of $\mathcal K$. $n_i$ is the leaf node where $n \rightarrow \text{\textit{text}} == \textit{``like''} $.\;
\For{$i = 0 \to m$}{
    $n = n_i \rightarrow \text{\textit{parent}}$.\;
    $c_v = $ \FMain{$n$}.\;
    $c_t = \texttt{NULL} $.\;
    \While{$n \neq  \texttt{NULL}$ \textbf{and} $c_t ==  \texttt{NULL}$}{
        \uIf {special rules for similes are triggered} {
            $c_t = $ \FMain{corresponding node}.\;
        } \ElseIf {$n \rightarrow $ \textit{label} in $\{\texttt{NP},\texttt{S}\}$} {
            $c_t = $ \FMain{$n$}.\;
        }
        $n = n \rightarrow \text{\textit{parent}}$.\;
    }
}

\KwRet \texttt{$c_t, c_v$}.\;
   \end{footnotesize}
   \end{algorithm}

\subsubsection{Property Generation}


A simile compares the topic $t$ to the vehicle $v$ via a shared property $p$~\cite{paul1970figurative}.
Hence, revealing the shared properties is crucial for promoting machines' ability to understand and generate similes.
Previous works skip properties acquisition since properties are usually implicit and can hardly be extracted from texts directly.

We generate properties for each simile sentence from two perspectives: \textit{knowledge} and \textit{context}.
Since \textit{vehicle} is the most valuable component when inferring the shared properties~\cite{tversky1977features}, 
we utilize the \textit{knowledge} from vehicles to retrieve the properties.
Specifically, given a vehicle, we generate properties via \textit{HasProperty} relation from COMET~\cite{bosselut2019comet}.
Besides, inferring the properties also requires an understanding of the \textit{context}~\cite{he2022can}.
Therefore, we rewrite collected simile sentence $s= ( w_1, ..., w_{i-1}, \textit{like}, w_{i+1}, ...,w_N )$ to ${s}'= ( w_1, ..., w_{i-1}, \textit{as}, \texttt{[MASK]}, \textit{as}, w_{i+1}, ...,w_N )$~\cite{he2022can}.
Then, we use $\text{RoBERTa}_{\text{LARGE}}$ with pre-trained masked-word-prediction heads to generate properties. 

We ensure the quality of the properties based on confidence scores.
We design dynamically adjusted thresholds rather than fixed ones.
For each simile sentence, we keep the top-10 predictions from each perspective, whose scores are normalized into [0, 0.5].
Only properties with a score greater than thresholds $\theta_{knowledge},\theta_{context}$ are retained. 

Overall, we obtain 5,511,111 \textit{simile instances} in the form of ($s$, $t$, $p$, $v$). 
From them, we extract 4,347,111 simile triplets.
For each ($s$, $t$, $p$, $v$), the final score is the sum of normalized scores predicted by COMET and $\text{RoBERTa}_{\text{LARGE}}$, which ranges from 0 to 1.
For each simile triplet, we calculate \textit{frequency} to present how many instances support it.

\subsubsection{A Probabilistic Knowledge Base}
\label{sec:probabilistic}
Previous works~\cite{li2013data} abandon extracted similes with low frequency.
Unpopular as the abandoned similes are, they may still be plausible and expressive.
Hence, instead of removing the uncertain similes, we model simile triplets with probabilistic information, which contains two metrics, \textit{plausibility} and \textit{typicality}~\cite{li2013data}.
\textit{Plausibility} evaluates the quality of simile triplets based on scores of their supporting instances, and 
\textit{typicality} measures how well a property suits a  topic-vehicle pair, and how typical it is to compare the topic to the vehicle given the property.

\vspace{2mm}
\textbf{Plausibility.} 
Each simile triplet ($t$, $p$, $v$) is supported by multiple ($s_{i}$, $t$, $p$, $v$) instances.
The quality of simile triplets depends on confidence scores of corresponding instances.
Hence, we adopt the \textit{noisy-or} model to measure \textit{plausibility} of simile triplets.
In the \textit{noisy-or} model, the plausibility of triplet ($t$, $p$, $v$) tends to be zero when confidence scores of all ($s_{i}$, $t$, $p$, $v$) instances are close to zero.
Formally, the plausibility of triplet ($t$, $p$, $v$) is defined as follows:
\begin{footnotesize}
\begin{gather*}
       \mathcal{P}(t,p,v) = 1 - \prod_{i = 1}^{\eta}(1 - S(s_i, t, p, v)),
\end{gather*}
\end{footnotesize}
where $S(s_i, t, p, v)=P(p|s_i, t, v)$ is the confidence score of each ($s_{i}$, $t$, $p$, $v$) instance in the previous step and 
$\eta$ is the number of instances supporting triplet ($t$, $p$, $v$).

\vspace{2mm}
\textbf{Typicality.}
Intuitively, \textit{curious} is a more typical property of (\textit{he},\textit{child}) compared with \textit{good}, and people will first think of  (\textit{he},\textit{child}) rather than  (\textit{he},\textit{cat}) given the property \textit{curious}.
Hence, the \textit{typicality} is an important measure for simile triplets, and can facilitate simile processing tasks.

We design two metrics to measure the \textit{typicality} of simile triplets: $\mathcal{T}(p|t,v)$ for simile understanding and $\mathcal{T}(t,v|p)$ for simile generation, which are formulated as follows:
\begin{footnotesize}
\begin{gather*}
        \mathcal{T}(p|t,v) = \frac{N(t,p,v)\cdot \mathcal{P}(t,p,v)}{\sum_{(t, {p}', v) \in \mathcal{G}_{(t,v)}} N(t,{p}',v)\cdot \mathcal{P}(t,{p}',v)},
        \\
        \mathcal{T}(t,v|p) = \frac{N(t,p,v)\cdot \mathcal{P}(t,p,v)}{\sum_{( {t}',p,{v}') \in \mathcal{G}_{p}} N({t}',p,{v}')\cdot \mathcal{P}({t}',p,{v}')}, 
\end{gather*}
\end{footnotesize}
where $\mathcal{G}_p$ or $\mathcal{G}_{(t, v)}$ denotes triplets containing $p$ or $(t, v)$ pair. $N(t, p, v)$ denotes the \textit{frequency} of $(t, p, v)$, i.e., the number of supporting instances $(s_{i}, t, p, v)$.
$\mathcal{P}(\cdot)$ are the plausibility. 


\section{MAPS-KB Statistics}
In this section, we compare the statistics of MAPS-KB with existing simile-related resources, as shown in Table \ref{tab:stat_resources}.
To begin with, MAPS-KB is million-scale,
while most of the others are hundred or thousand-scale.
Although there is one million-scale knowledge base MetaKB~\cite{li2013data}, it lacks the important component \textit{property}, which plays an imperative role in downstream tasks.
Moreover, MAPS-KB is the only probabilistic simile knowledge base.

\begin{table}[h] 
    \small
    \resizebox{0.48\textwidth}{12mm}{
    \centering
        \begin{tabular}{c|c|c|c|c|c}
        \toprule  
        \textbf{Resource}   & \textbf{Size}   & \textbf{\#$t$} & \textbf{\#$v$} & \textbf{\#$p$ } & \textbf{\textit{Prob}}\\
        \midrule  
        Linguistics~\cite{roncero2015semantic}   & 679 & 65 & 75 & 379 & N\\
        SPP~\cite{he2022can}  & 1633 & 721 & 667 & 333  & N\\
        Fig-QA(M)~\cite{liu2022testing}  & 1458 & 441 & 1198 & 646 & N\\
        MetaKB~\cite{li2013data}   &   2.6M   &    -  &  -   &  -     & N  \\
        MetaKB*~\cite{li2013data}  &   0.9M  &    383k  &  434k   &  -    & N  \\
        \midrule
        MAPS-KB  &    4.3M  &   178k   &   227k  &     9k   & Y \\
        \bottomrule 
        \end{tabular}
    }
        \captionsetup{font={small}} 
        \caption{The statistics of MAPS-KB and existing simile-related resources. MetaKB* denotes the published subset of MetaKB. Fig-QA(M) denotes the medium training set which reports the statistics. \textit{Prob} indicates whether probabilistic information is contained.}
    \label{tab:stat_resources}
\end{table}

To further compare with the only existing million-scale knowledge base MetaKB, we show the frequency distribution of ($t$, $v$) pairs in Figure \ref{fig:distribution} (left).
Although both MAPS-KB and MetaKB* exhibit long-tail distributions, the tail of MetaKB*'s distribution is much heavier than that of MAPS-KB's distribution.
This indicates that MAPS-KB contains more reliable ($t$, $v$) pairs which appear multiple times.


According to the conceptual metaphor theory~\cite{lakoff1993contemporary}, metaphor can be viewed as a mapping between terms in two domains, which reflects people's understanding of the world.
Hence, we study the domain mapping patterns of similes by distinguishing topic and vehicle into ten domains via WordNet~\cite{miller1995wordnet} hypernyms.
We select ten domains: \{ \textit{person}, \textit{animal}, \textit{body part}, \textit{food}, \textit{natural object}, \textit{natural phenomenon}, \textit{feeling}, \textit{artifact}, \textit{location}, \textit{action}\}. 
The method of selecting domains and assigning them to terms is detailed in Appendix.
Figure~\ref{fig:categorization} shows the domain distribution of topics and vehicles. 
We also show the examples and percentages of the top 5 domain mappings between topics and vehicles in Table \ref{tab:mapping_statistics}.

\begin{table}[t] 
\small
    \centering
        \begin{tabular}{ccc}
        \toprule  
        $(c_t, c_v)$ \textbf{Pair} & \textbf{Example Simile Triplet}  & $\%$\\ 
        \midrule  
            (artifact, artifact) & (road, decorative, ribbon) & 10.61 \\
            (person, person) & (man, funny, fool) & 10.05 \\
            (person, artifact) & (man, silent, statue) & 06.24  \\
            (location, artifact) & (studio, complex, spaceship)  & 04.86 \\
            (artifact, person) & (large cloak, devout, priest) & 03.79  \\
        
        \bottomrule 
        \end{tabular}
        \caption{Examples and percentages of the top 5 domain mappings between topics and vehicles. For each domain mapping, \% denotes the percentage of corresponding simile triplets among all triplets.}
    \label{tab:mapping_statistics}
\end{table}

\begin{figure}[t]
    \centering
        \includegraphics[width=0.8\linewidth]{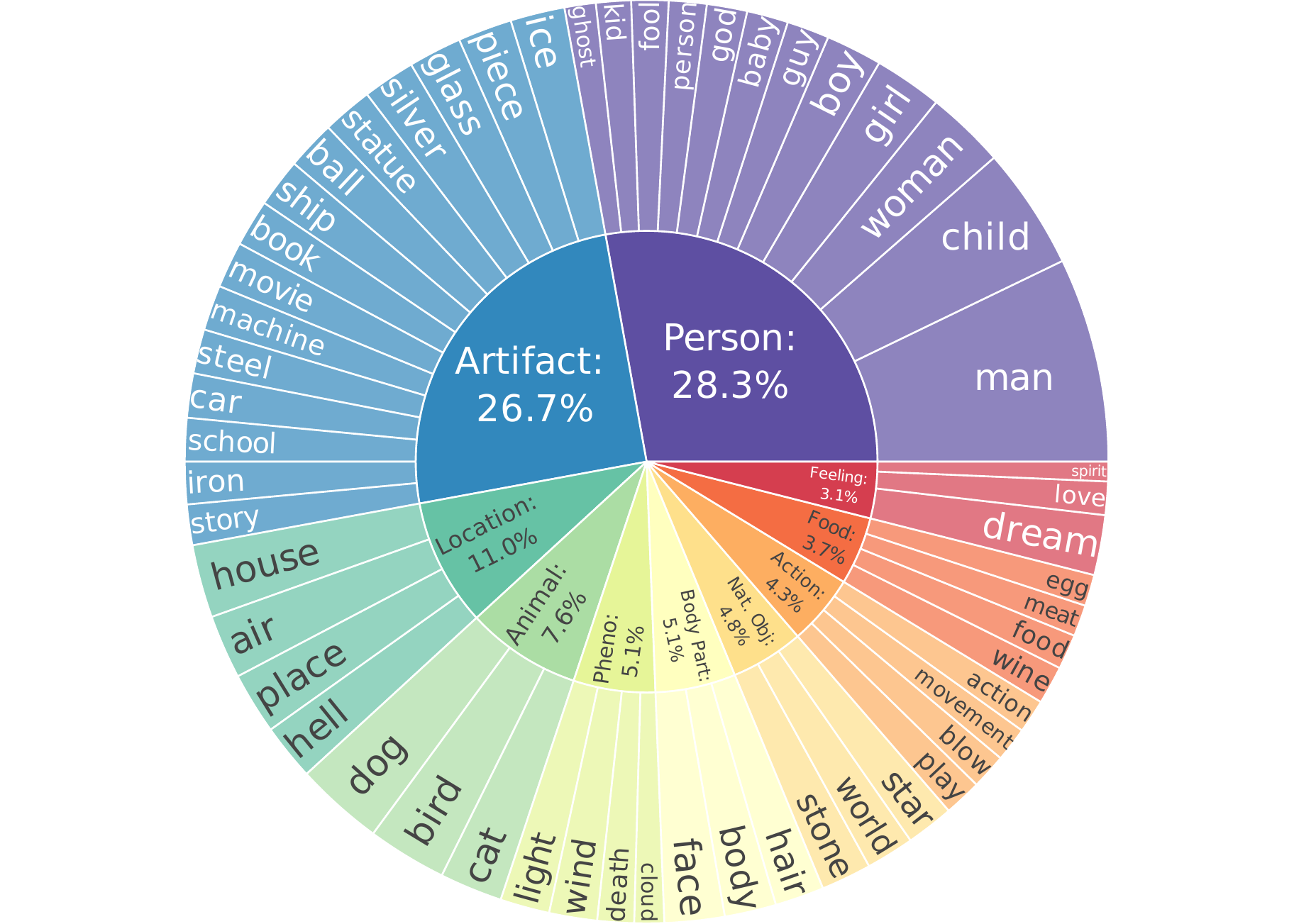} 
    \captionsetup{font={small}} 
    \caption{Distribution of categories of topics and vehicles. Topics and vehicles are assigned to 10 categories via WordNet hypernyms.}
    \label{fig:categorization}
\end{figure}

\begin{figure}[t]
    \centering
        \includegraphics[width=1\linewidth]{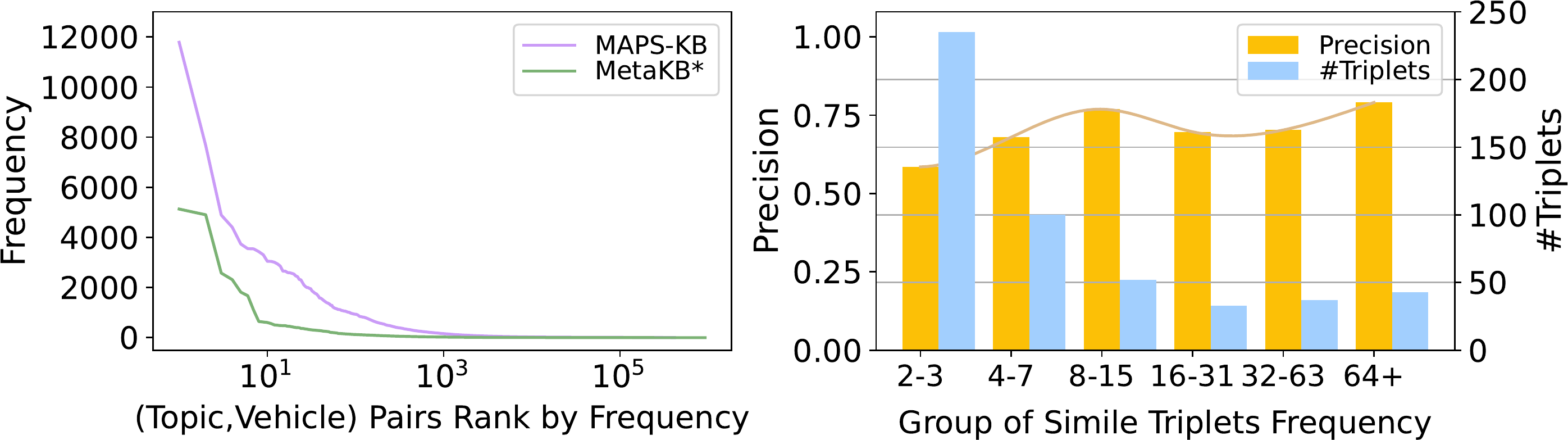}
    \captionsetup{font={small}} 
    \caption{The frequency distribution of the ($t$, $v$) pairs  in MAPS-KB and MetaKB* (left), and human evaluation results of simile triplets in MAPS-KB grouped by frequency (right). }
    \label{fig:distribution}
\end{figure}


\section{Intrinsic Evaluation}
In this section, we conduct experiments to evaluate the effectiveness and necessity of methods in our framework as well as the quality of simile triplets in MAPS-KB.

\paragraph*{Simile Detection}
We evaluate the effectiveness of our simile detection method.
We collect 1,197 sentences of \textit{like} view from Wikipedia and ask two annotators to label them.
As a result, 401 similes and 796 literal sentences are labeled.
We also consider existing labeled dataset UGC~\cite{niculae2014brighter}, which contains 368 similes and 405 literal sentences of \textit{like} view.
The sentences are split into 8:1:1 as the train/dev/test splits.


We compare the performance of  $\text{BERT}_{\text{BASE}}$ classifier(s) in three settings: 
(1) Supervision: The model is trained entirely on labeled data.
(2) Self-supervision: Besides labeled data, we also utilize unlabeled data from \textit{like} view.
(3) Co-training: We leverage unlabeled data from \textit{like} view and \textit{be} view, as is introduced in the previous section.

The results are shown in Table \ref{tab:detection_result}.
The results of all our experiments are averaged over three random seeds. According to the results, we find that
co-training achieves the best precision and F1 score with comparable recall, which demonstrates that co-training can detect most of the simile sentences while ensuring better quality of the predicted ones.
As we concern more about precision in this step, co-training is a better solution than other methods.


\begin{table}[]
    \centering
    \resizebox{0.46\textwidth}{10.5mm}{
\begin{tabular}{c|ccc|ccc}
        \toprule  
\textbf{Dataset }         & \multicolumn{3}{c|}{\textbf{Wikipedia}} & \multicolumn{3}{c}{\textbf{UGC}} \\
        \midrule  
\textbf{Metric}           & \textbf{P}        & \textbf{R}        & \textbf{F1}      & \textbf{P}          & \textbf{R }        & \textbf{F1}        \\
        \midrule  
Supervision      & 71.18    & 78.33    & 74.58   & 81.73      & 77.16     & 79.50     \\
Self-supervision & 73.38    & \textbf{91.67}    & 81.49   & 83.20      & \textbf{78.33}     & 80.66     \\
Co-training      & \textbf{74.50}    & 90.00    & \textbf{81.52}   & \textbf{84.67}      & 77.50     & \textbf{80.88}    \\
        \bottomrule 
\end{tabular}
}
    \caption{Results of different training methods in  simile detection.}
    \label{tab:detection_result}
\end{table}

 
\paragraph*{Component Extraction}
We also evaluate the quality of our component extraction method.
Two annotators are asked to label the components of similes in the previous section.
Annotated components should cover rich semantics as much as possible.
For example, in ``\textit{He is like a kid in the candy store.}'', it is better for the vehicle to be ``\textit{kid in the candy store}'' rather than simply ``\textit{kid}''.
We finally get 401 similes with labeled components. 
Existing labeled datasets~\cite{niculae2014brighter} are not concerned, since the labeled components are only one word.

We compare components extracted by the rules of MetaKB and ours. 
MetaKB designs rules solely based on regex, without considering syntactic structure.
For evaluation, we consider two settings with different difficulties. 
In \textit{hard} setting, a prediction is considered correct only if it matches the label exactly, while in
\textit{easy} setting, a prediction is simply expected to have the same core noun as the label.
From the results in Table \ref{tab:component_result}, we have the following analyses.
First of all, our rules outperform those of MetaKB significantly under both settings.
MetaKB rules are only comparable to ours for vehicles in the  \textit{easy} setting.
This indicates that syntactic structure plays an imperative role in component extraction.
Second, extracting topics is more difficult than extracting vehicles.
This is because topics tend to have longer distance from the comparator ``\textit{like}'' than vehicles.
Additionally, with regards to vehicles, the performance gap between rules of ours and MetaKB is much more significant in the \textit{hard} setting than \textit{easy}.
This indicates that though regex-based rules can locate vehicles correctly, it is hard for them to identify the integral vehicles with full semantics.

\begin{table}[]
\small
\centering
    \resizebox{0.40\textwidth}{19mm}{
\begin{tabular}{ccllll}
    \toprule
\textbf{Eval}     & \textbf{Comp}       & \multicolumn{1}{c}{\textbf{Method}} & \multicolumn{1}{c}{\textbf{P}} & \multicolumn{1}{c}{\textbf{R}} & \multicolumn{1}{c}{\textbf{F1}} \\
         \midrule
\multirow{4}{*}{Hard} & \multirow{2}{*}{Topic}   & MetaKB                              & 53.28                                  & 51.13                               & 52.19                           \\
                      &                          & MAPS-KB                               & \textbf{64.50}                         & \textbf{59.95}                      & \textbf{62.14}                  \\
                        \cmidrule{2-6}
                      & \multirow{2}{*}{Vehicle} & MetaKB                              & 67.10                                  & 64.41                               & 65.73                           \\
                      &                          & MAPS-KB                               & \textbf{81.79}                         & \textbf{77.69}                      & \textbf{79.69}                  \\
                               \midrule
\multirow{4}{*}{Easy} & \multirow{2}{*}{Topic}   & MetaKB                              & 71.39                                  & 68.51                               & 69.92                           \\
                      &                          & MAPS-KB                               & \textbf{81.30}                         & \textbf{76.92}                      & \textbf{79.05}                  \\
                        \cmidrule{2-6}
                      & \multirow{2}{*}{Vehicle} & MetaKB                              & \textbf{92.43}                         & \textbf{88.72}                      & \textbf{90.54}                  \\
                      &                          & MAPS-KB                               & 92.35                                  & 88.38                               & 90.32            \\              
                          \bottomrule
\end{tabular}
}
    \caption{Results of rules of MetaKB and ours in the component extraction task. \textit{Hard} and \textit{Easy} represent different levels of difficulty.}
    \label{tab:component_result}
\end{table}

\subsection{Human Evaluation}
We conduct human evaluation to evaluate the precision of simile triplets in MAPS-KB. 
We randomly sample 500 triplets with frequency greater than one, and ask two annotators to label whether they are correct or not. 
The average precision is 0.71, and the Fleiss' Kappa score~\cite{fleiss1971measuring} is 0.75.
We further study the precision of triplets grouped by frequency.
According to the results shown in Figure \ref{fig:distribution} (right),
we find that simile triplets with higher frequency tend to be more precise, which is a reasonable result.

\section{Extrinsic Evaluation}
In this section, we apply MAPS-KB knowledge to three downstream applications to demonstrate its effectiveness.

\subsection{MAPS-KB for Simile Processing Tasks}
There are mainly two simile processing tasks: simile interpretation (SI) and simile generation (SG).
SI aims to infer shared properties for given ($t$, $v$) pairs, while SG aims to generate suitable vehicles for given ($t$, $p$) pairs.

\subsubsection{Inference Rules}
We define scores $S_{(t, v)}(p)$ and $S_{(t, p)}(v)$ based on \textit{typicality}, \textit{frequency} and \textit{plausibility} for the inference of SI and SG respectively.
The candidate properties and vehicles are ranked by corresponding scores, and the ones with higher scores are regarded as possible answers. These scores are formulated as follows:
\begin{footnotesize}
\begin{gather*}
\mathcal S_{(t, v)}(p) =\sum_{({t}', p, v) \in \mathcal{G}_{(p,v) }} \mathcal T(p|{t}',v) \cdot N({t}',p,v) \cdot \mathcal{P}({t}',p,v),\\
\mathcal S_{(t, p)}(v) = \sum_{({t}', p, v) \in \mathcal{G}_{(p,v) }} \mathcal T({t}',v|p) \cdot N({t}',p,v) \cdot \mathcal{P}({t}',p,v).
\end{gather*}
\end{footnotesize}
Specially, we ignore the original topic $t$ and sum over all topics ${t}'$ of $({t}',p,v) \in \mathcal{G}_{(p,v) }$ that serve as topics of $(p, v)$ in MAPS-KB. The reasons are twofold:
(1) 
Theoretically, properties and vehicles are more salient to each other compared with topics~\cite{tversky1977features, veale2007learning}.
(2) Practically, 
MAPS-KB cannot cover the probabilistic information of all $(t, p, v)$ triplets in downstream datasets, which hinders inference. 
Considering both reasons, we adopt the more available statistic information of $(p, v)$.



For both tasks, the scores are sums of products of  \textit{typicality} $\mathcal T$, \textit{frequency} $N$ and \textit{plausibility} $\mathcal P$, which empirically yield the best results compared with the alternatives that concern only one or two of them. 


\subsubsection{Experiment Setup}
We evaluate our inference rules on three benchmark datasets: Linguistics~\cite{roncero2015semantic}, Quizzes and General Corpus~\cite{he2022can}. 
Following~\cite{chen2022probing}, we keep the similes in Linguistics with frequency larger than 4. 
We use Mean Reciprocal Rank (\textbf{MRR}) and Recall@k (\textbf{R@k}) as metrics.

We compare our inference rules with the following baselines:
(1) \textbf{Meta4meaning}~\cite{xiao2016meta4meaning}:
They adopt statistical associations to select optimal properties.
(2) \textbf{GEM}~\cite{bar2020automatic}:
They rank properties via their co-occurrence and similarity with topics and vehicles.
(3) \textbf{ConScore}~\cite{zheng2019love}:
They rank candidates via their distance from other components in embedding space.
(4) \textbf{LMASKB}~\cite{chen2022probing}: 
They probe simile knowledge from PLMs via designed prompt templates.
Note that  {ConScore} and {LMASKB} utilize the train split, while {Meta4meaning}, {GEM} and our method do not.

\subsubsection{Results}

The results are shown in Table \ref{tab:KBcompletion_result}, from which we draw the following conclusions.
First, MAPS-KB with our inference rules achieves the best performance on most of the metrics for both tasks. 
Even though our method does not utilize the train set, it surpasses supervised methods ConScore and LMASKB which do.
This validates that MAPS-KB contains abundant informative simile knowledge that well supports downstream simile processing tasks.
Besides, our method outperforms LMASKB which directly probes simile knowledge from PLMs.
This reveals that our explicit structure knowledge is more useful than the implicit statistical knowledge contained in PLMs when processing similes.

\begin{table*}[t]
    
        \resizebox{1\textwidth}{19mm}{
\begin{tabular}{ccc|ccccc|ccccc|ccccc}
    \toprule
\multicolumn{3}{c}{\textbf{Dataset}} & \multicolumn{5}{c}{\textbf{Linguistics}}     & \multicolumn{5}{c}{\textbf{Quizzes}} & \multicolumn{5}{c}{\textbf{General Corpus}} \\
          \midrule
\multicolumn{1}{l}{\textbf{Task}} & \textbf{Method } & \textbf{TS}      & \textbf{MRR}   & \textbf{R@5}  & \textbf{R@10}  & \textbf{R@15}  & \textbf{R@25}  & \textbf{MRR}   & \textbf{R@5}  & \textbf{R@10}  & \textbf{R@15}  & \textbf{R@25} & \textbf{MRR}   & \textbf{R@5}  & \textbf{R@10}  & \textbf{R@15}  & \textbf{R@25}   \\
           \midrule
\multirow{6}{*}{\textbf{SI}}      & Meta4meaning & N & N/A   & 0.221 & 0.303 & 0.339 & 0.397  & 0.153   & 0.222 & 0.276 & 0.314 & 0.465 & 0.069  & 0.094 & 0.156 & 0.195 & 0.247  \\
                         & GEM & N          & \underline{0.312} & 0.198 & 0.254 & 0.278 & 0.405 & 0.075 & 0.055 & 0.116 & 0.154 & 0.233 & 0.030 & 0.023 & 0.055 & 0.075 & 0.106 \\
                         & ConScore & Y     & 0.078 & 0.076 & 0.138 & 0.172 & 0.269 & 0.044 & 0.048 & 0.112 & 0.167 & 0.229 & 0.030 & 0.029 & 0.052 & 0.084 & 0.131 \\
                         & LMASKB & Y    & 0.270 & \textbf{0.378} & \underline{0.490} & \underline{0.524} & \underline{0.579}   & \underline{0.342} & \underline{0.440} & \underline{0.549} & \underline{0.628} & \underline{0.716}  & \underline{0.196} & \underline{0.271} & \underline{0.352}& \underline{0.419} & \underline{0.522}  \\
                           \cmidrule{2-18}
                         & MAPS-KB & N        & \textbf{0.392 }& \underline{0.367} & \textbf{0.493} & \textbf{0.544} & \textbf{0.587}   & \textbf{0.375 }& \textbf{0.520} & \textbf{0.699} & \textbf{0.749} & \textbf{0.806}  & \textbf{0.200 }& \textbf{0.318} & \textbf{0.439} & \textbf{0.493} & \textbf{0.553}  \\
                          \midrule
\multirow{4}{*}{\textbf{SG}}      & ConScore & Y     & 0.036 & 0.055 & 0.090 & 0.103 & 0.145  & 0.002 & 0.002 & 0.007& 0.007 & 0.009 & 0.001 & 0.001 & 0.003 & 0.004& 0.010 \\
                         & LMASKB & Y    & \underline{0.095} & \underline{0.124} & \underline{0.145} & \underline{0.159} & \underline{0.214}  & \underline{0.042} & \underline{0.059} & \underline{0.094} & \underline{0.121} & \underline{0.154} & \underline{0.022} & \underline{0.026} & \underline{0.037} & \underline{0.046} & \underline{0.067}  \\
                         \cmidrule{2-18}
                         & MAPS-KB & N        & \textbf{0.105} & \textbf{0.140} & \textbf{0.162} & \textbf{0.179} & \textbf{0.217}     & \textbf{0.132} & \textbf{0.201} & \textbf{0.272} & \textbf{0.336} & \textbf{0.423}    & \textbf{0.067} & \textbf{0.092} & \textbf{0.115} & \textbf{0.144} & \textbf{0.168}\\
        
    \bottomrule
    \end{tabular}
    }
    \caption{Evaluation results of different methods on simile interpretation(SI) task and simile generation(SG) task. Bold numbers are the best results. The second best results are marked by ``$\underline{\ \ \ \ \ \ }$''. 
    TS indicates whether the train set is used. 
    On the Linguistics dataset, results of ConScore and LMASKB are taken from~\cite{chen2022probing}, other results are taken from their original papers. }
    \label{tab:KBcompletion_result}
\end{table*}

\subsection{MAPS-KB for Writing Polishment}
We propose three methods to polish writing with MAPS-KB: a PLM-based method, a rule-based method, and their combination.
The PLM-based method first injects simile knowledge in MAPS-KB into a sequence-to-sequence PLM BART~\cite{lewis2019bart} via finetuning.
The finetuning dataset is automatically constructed from MAPS-KB, where 
a collected simile instance $(s,t,p,v)$ is transformed into a sample, whose output decoder target is the simile sentence $s$ itself and the input encoder source is $s$ rewritten to drop $v$ but include $p$. 
For example, given $s=$``\textit{Her hair felt like silk.}'' and $(t, p, v)=$(\textit{hair}, \textit{soft}, \textit{silk}), the encoder source would be ``\textit{Her hair is soft.}''.
Afterward, the PLM can be directly applied to downstream datasets without further training. 

The rule-based method infers vehicles from probabilistic information of MAPS-KB, and then rewrites sentences with rules. 
Given the adjective/adverb property $p$ in the input, the score $S'_p(v)$ for vehicle $v$ is defined as follows:
\begin{footnotesize}
\begin{gather*}
\mathcal S'_p(v)=\sum_{({t}',p,v)\in \mathcal{G}_{(p,v) }} \mathcal T({t}',v|p) \cdot N({t}',p,v)  \cdot \mathcal{P}({t}',p,v) \cdot e^{\gamma \cdot l(v)}.
\end{gather*}
\end{footnotesize}

Compared with $S_p(v)$ in simile processing, $S'_p(v)$ has an additional term $e^{\gamma \cdot l(v)}$, where $l(v)$ is the number of words in $v$ and $\gamma$ is a hyperparameter.
By introducing $e^{\gamma \cdot l(v)}$, we encourage longer vehicles which tend to be more expressive. 
Then, we replace the adjective/adverb $p$ in the input with a comparator ``\textit{like}'' and the vehicle $v$ of highest $S'_p(v)$ score.

Finally, we propose a combination of the PLM-based and rule-based methods. 
We find that PLM-generated similes may be incoherent if (1) they do not contain ``\textit{like}'' or (2) their vehicles contain commas or are longer than seven words. 
The combined method outputs similes generated by rules in this case, and similes generated by PLMs otherwise. 


\subsubsection{Experiment Setup}
We evaluate our methods on the test set proposed by~\cite{chakrabarty2020generating}.
In this testset, the literal inputs always end with adjective/adverb, while the simile outputs end with ``\textit{like}'' and vehicles.
E.g., the literal input ``\textit{Love is rare}'' is expected to be polished into the simile output ``\textit{Love is like a unicorn}''.

For evaluation, we only retain the generated vehicles after the word ``\textit{like}'', following~\cite{chakrabarty2020generating}.
We adopt three automatic evaluation metrics:
(1) \textbf{BLEU}~\cite{papineni2002bleu}, one of the most popular metrics for assessing generation tasks.
(2) \textbf{BERTScore}~\cite{zhang2019bertscore}, which measures semantic similarity with BERT.
(3) \textbf{Average Length} of generated vehicles, which is important because longer vehicles are generally more expressive and contain more semantics.  


We compare our methods with following baselines:
(1) \textbf{RTRVL}~\cite{chakrabarty2020generating}: 
They adopt the HasProperty relation from ConceptNet~\cite{speer2017conceptnet} to choose the optimal vehicle.
(2) \textbf{SCOPE}~\cite{chakrabarty2020generating}:
They finetune BART via automatically constructed parallel datasets.
(3) \textbf{META}~\cite{stowe2020metaphoric}:
They finetune BART via parallel data containing masked literal sentence and similes.
(4) \textbf{LMASKB}~\cite{chen2022probing}:
They generate vehicle by prompting via designed templates.
(5) \textbf{GPT-3}~\cite{brown2020language}: We prompt GPT3-Davinci-002 to generate similes given 4 random demonstrative examples.

\begin{table}[t] 
    \small
    \centering
    \resizebox{0.48\textwidth}{16.8mm}{
        \begin{tabular}{c|c|c|c|c|c}
        \toprule  
       \textbf{ Method  } & \textbf{TS}    & \textbf{BLEU1} & \textbf{BLEU2 }& \textbf{BERT-S}& \textbf{Length} \\
        \midrule  
        RTRVL & N &       00.00       &  00.00       &  12.94          &  01.59  \\
        LMASKB   & N &   00.53      &   00.00   &     14.05       &  01.00 \\
        META   & Y &    03.73      &   00.96      &   15.14              &  01.58  \\
        SCOPE   & Y &     08.03        &     \textbf{03.59}    &  \underline{18.04}              &  01.87  \\
        GPT-3   & N &     \underline{12.40}        &     02.87    &  16.96              & \textbf{03.20}  \\
    \midrule  
        $\text{MAPS-KB}_{\textit{PLM}}$   & N &     11.99       &   03.13   &            16.65      &  \underline{02.95} \\
        $\text{MAPS-KB}_{\textit{Rule}}$   & N &    04.80       &  03.27   &   10.47            & 02.03 \\
        $\text{MAPS-KB}_{\textit{PLM+Rule}}$   & N &     \textbf{13.11 }       &     \underline{03.44}    &  \textbf{18.94}            &  02.71   \\
        \bottomrule 
        \end{tabular}
        }
        \captionsetup{font={small}} 
        \caption{Results of metrics BLEU-1, BLEU-2, BERTScore and Average Length in the writing polishment task.
        TS indicates whether the train set from ~\cite{chakrabarty2020generating} is used.
        Results of RTRVL, META and SCOPE are taken from~\cite{chakrabarty2020generating}. }
        
    \label{tab:writing_polishmen}
\end{table}

\subsubsection{Results}
Results shown in Table \ref{tab:writing_polishmen} suggest that our methods significantly outperform prior methods in this task, even though we do not use the train set. 
This not only validates the effectiveness and quality of simile knowledge in MAPS-KB, but also 
indicates that our methods with MAPS-KB can rewrite literal sentences to similes that are more plausible (higher BLEU1, BLEU2 and BERT-S) and contain more semantics (higher Average Length).
Besides, the combined method $\text{MAPS-KB}_{\textit{PLM+Rule}}$ performs better than methods based solely on PLMs or rules. 
The improvement of $\text{MAPS-KB}_{\textit{PLM+Rule}}$ over $\text{MAPS-KB}_{\textit{PLM}}$ shows
the practical value of MAPS-KB' probabilistic information in this task.  
Furthermore, $\text{MAPS-KB}_{\textit{Rule}}$ 
surpasses the retrieval method RTRVL based on ConceptNet, which implies that MAPS-KB is better than existing commonsense knowledge bases in terms of polishing writing with similes.

\section{Conclusions}
In this work, we introduce MAPS-KB, the first million-scale probabilistic simile knowledge base. 
Specifically, MAPS-KB covers 4.3 million triplets with probabilistic information over 0.4 million terms from 70 GB corpora.
The proposed construction framework consists of four steps and can be extended to other figurative languages.
We further conduct intrinsic and extrinsic evaluation to verify the effectiveness of our KB and framework, and achieve state-of-the-art performance on three downstream tasks.

\section*{Acknowledgements}
This work is supported by Shanghai Municipal Science and Technology Major Project (No.2021SHZDZX0103), Science and Technology Commission of Shanghai Municipality Grant (No. 22511105902), National Natural Science Foundation of China (No.62102095).
Yanghua Xiao is also a member of Research Group of Computational and AI Communication at Institute for Global Communications and Integrated Media, Fudan University.

\bibliography{aaai23}
\clearpage

\section{Appendix}
\subsection{Special Cases for Components Extraction}
In the components extraction section, when iterating over the parent nodes of $n_i$ from bottom to top (line 15-20 in Algorithm 2), the \textit{topic} will appear in the subtree rooted at the special node if some special cases for similes are triggered (line 16-17). 
Traversing the parent nodes of $n_i$ from bottom to top, we get a sequence constituted by the label of each parent node. 
For example, in the second step of Figure 1 of the main paper, when traversing the parent nodes of $n_1$, the label sequence is [\texttt{PP}, \texttt{VP}, \texttt{VP}, \texttt{S}].
The cases are as follows:
\begin{enumerate}
    \item If the label sequence contains the subsequence [\texttt{NP}, \texttt{SBAR}, \texttt{S}], the topic will appear in the subtree rooted at \texttt{S} rather than \texttt{NP}. 
    The example for this case is the attributive clause (e.g. In m - system engines, the fuel is injected onto the walls of \textit{the combustion chamber}, that is solely located inside the piston, and shaped \textit{like} \textit{a sphere}.).
    \item The label sequence contains the subsequence [$\texttt{VP}_1$, $\texttt{VP}_2$, $\texttt{VP}_3$], the subsequence appears before \texttt{S} or \texttt{NP} and the text of $\texttt{VP}_2$ is ``to''. 
    Here, the subscripts denote the order in which the label \texttt{VP} appears. 
    The topic may appear in the subtree rooted at the $\texttt{VP}_3$.
    The example for this case is the specific verb phrase (e.g. Alessandro tells \textit{Aminta} to dress \textit{like} \textit{a king} so he can be presented to his subjects.).
    \item If the extracted \textit{topic} is a pronoun (e.g. \textit{it}, \textit{that}, \textit{them}), we replace the pronoun with its referent via CoreNLP~\footnote{https://stanfordnlp.github.io/CoreNLP/}. (e.g. \textit{His paintings} are executed with a precision that makes them look \textit{like} \textit{photographs}.).
\end{enumerate}

\subsection{Filter Rules Based on Extracted Components}
After components extraction, we further filter simile sentences based on extracted components due to the noise in the simile detection process.
The filter rules are as follows:
\begin{enumerate}
    \item Remove the sentence if the topic or vehicle is a gerund, since it is an analogy rather than a simile. (e.g. \textit{Creating Artificial Intelligence} is \textit{like} \textit{summoning the demon}.).
    \item Remove the sentence if the nouns in topic and vehicle overlap, since the sentence is likely to be a literal comparison rather than a simile. (e.g. \textit{His death} must be \textit{like} \textit{all other human death}.).
    \item Remove the sentence if the topic is a pronoun except personal pronoun (e.g. \textit{it}, \textit{that}, \textit{something}), since the topic does not contain useful semantic information. (e.g. \textit{It} was \textit{like} \textit{a dream} , I babbled , more to myself than Selena.).
\end{enumerate}

\subsection{Domain Mapping Patterns of Similes}
In the MAPS-KB Statistics section of the main paper, we study the domain mapping patterns of similes in MAPS-KB.
To select domains, we collect WordNet hypernyms of terms (i.e. topic and vehicle), and manually select ten popular and general hypernyms as the domains that cover most of the terms:  \{ \textit{person}, \textit{animal}, \textit{body part}, \textit{food}, \textit{natural object}, \textit{natural phenomenon}, \textit{feeling}, \textit{artifact}, \textit{location}, \textit{action}\}.

To assign domains to terms, we select nouns in terms via part-of-speech tagging using NLTK~\footnote{https://www.nltk.org/}.  
Then, for a given noun, we traverse its synsets until finding a domain in the hypernym path of the synset, defined as the domain of the term. 
Long or uncommon terms are ignored if they are not recognized by NLTK part-of-speech tagging, not found in WordNet,  or not assigned to selected domains.
We consider the frequency of simile triplets as their weights in distribution calculation.
Table~\ref{tab:category_statistics} shows the examples and percentages of topics and vehicles in different domains. 

\begin{table}[h] 
\small
    \centering
        \begin{tabular}{cccc}
        \toprule  
        \textbf{Category} & \textbf{Example} & $\%_t$ & $\%_v$  \\ 
        \midrule  
        Person & child & 25.98 & 29.55 \\
        Body Part & cheeks & 08.57 & 03.23\\
        Animal & dog & 03.44 & 10.27 \\
        Food & chicken & 02.96 & 04.11 \\
        Feeling & grief & 04.08 & 02.57  \\
        Action & blow & 05.52 & 03.70  \\
        Natural Phenomenon & rain & 04.92 & 05.24  \\
        Natural Object & stone & 04.29 & 05.03 \\
        Artifact & statue & 25.65 & 27.21  \\
        Location & home & 14.60 & 09.09 \\
        \bottomrule 
        \end{tabular}
        \caption{Examples and percentages of topics and vehicles in different domains. Here, $t$ and $v$ denote topic and vehicle respectively.}
    \label{tab:category_statistics}
\end{table}

\subsection{Experimental Details}
All the experiments run on RTX3090 GPU.
The implementations of all the PLMs are based on the HuggingFace Transformers~\footnote{https://github.com/huggingface/transformers/}.
During simile detection, the experiments are run with batch sizes of 64, max sequence length of 128, and learning rate of 4e-5 for 50 epochs. 
All the hyper-parameter settings are shown in Table~\ref{tab:parameter}.

\begin{table}[h] 
\footnotesize
    \centering
      \resizebox{0.48\textwidth}{32mm}{
        \begin{tabular}{ccc}
        \toprule  
        \textbf{Notation} & \textbf{Description} & \textbf{Setting}    \\ 
        \midrule  
            $\alpha_{like}$ & \begin{tabular}[c]{@{}l@{}}Sample ratio of unlabeled data from \textit{like} \\  view in simile detection task.\end{tabular} & 0.10\% \\
            \midrule
            $\alpha_{be}$ & \begin{tabular}[c]{@{}l@{}}Sample ratio of unlabeled data from \textit{be}  \\  view in simile detection task.\end{tabular} & 0.01\% \\
            \midrule
            $\theta_{like}$ & \begin{tabular}[c]{@{}l@{}}Confidence score threshold from \textit{like} \\  view in simile detection task.\end{tabular} & 0.9 \\
            \midrule
            $\theta_{be}$ & \begin{tabular}[c]{@{}l@{}}Confidence score threshold from \textit{be}  \\  view in simile detection task.\end{tabular} & 0.9 \\
            \midrule
            $T$ & Best iteration times in simile detection task. & 5 \\
            \midrule
            $\theta_{knowledge}$ & \begin{tabular}[c]{@{}l@{}}Confidence score threshold from \textit{knowledge} \\ perspective in property generation task.\end{tabular}& 0.3  \\
        \midrule
            $\theta_{context}$ & \begin{tabular}[c]{@{}l@{}}Confidence score threshold from \textit{context} \\ perspective in property generation task.\end{tabular}& 0.0 \\
        \midrule
          $\gamma$ & \begin{tabular}[c]{@{}l@{}}The effect of vehicle length on the final score \\ for writing polishment task. \end{tabular} & 2  \\
            
        \bottomrule 
        \end{tabular}
        }
        \caption{The description and setting of important hyper-parameters.}
    \label{tab:parameter}
\end{table}


\end{document}